\documentclass{article}
\usepackage{ijcai20}

\usepackage{algorithm}
\usepackage{algorithmicx}
\usepackage{algpseudocode}
\usepackage{amsfonts}
\usepackage{times}

\usepackage{soul}
\usepackage{url}
\usepackage[utf8]{inputenc}
\usepackage[small]{caption}
\usepackage{graphicx}
\usepackage{amsmath}
\usepackage{booktabs}
\usepackage{graphicx}
\usepackage{subfigure}
\usepackage{authblk}
\usepackage[normalem]{ulem}
\useunder{\uline}{\ul}{}

\title{Explainable Deep Convolutional Candlestick Learner}

\author[1]{Jun-Hao Chen}
\author[2]{Samuel Yen-Chi Chen}
\author[3*]{Yun-Cheng Tsai}
\author[4]{Chih-Shiang Shur}
\affil[1]{Department of Computer Science and Information Engineering, National Taiwan University}
\affil[2]{Department of Physics, National Taiwan University}
\affil[3]{School of Big Data Management, Soochow University}
\affil[4]{FinTech Center, Soochow University}
\affil[*]{Corresponding Email: pecutsai@gm.scu.edu.tw}
\begin{document}

\maketitle

\begin{abstract}
Candlesticks are graphical representations of price movements for a given period. The traders can discovery the trend of the asset by looking at the candlestick patterns. Although deep convolutional neural networks have achieved great success for recognizing the candlestick patterns, their reasoning hides inside a black box. The traders cannot make sure what the model has learned. In this contribution, we provide a framework which is to explain the reasoning of the learned model determining the specific candlestick patterns of time series. Based on the local search adversarial attacks, we show that the learned model perceives the pattern of the candlesticks in a way similar to the human trader.

Keywords: local search adversarial attacks, explainable artificial intelligence, candlesticks, time series encoding, convolutional neural network, financial vision.

\end{abstract}

\section{Introduction}
The candlestick patterns recognition lies at the heart of trading and the foundation of all technical analysis.
Therefore, understanding how to interpret candlestick is a critical step in becoming a trader.
Hence, traders need a candlestick patterns recognition tool to help them discover valuable information from candlestick.
Although object detection and pattern recognition technologies have been prevailed in the computer vision field, traders generally cannot rely on these tools to gain insights of the candlestick patterns due to the lack of acquiring trading knowledge-based feature representations. 

According to Tsai et al., they proposed an extended Convolutional Neural Networks (CNN) approach to recognize the candlestick patterns automatically~\cite{tsai2019encoding}. Even though the deep learning based model has three significant advantages, including non-linearity, robustness, and adaptive manner, the traders cannot trust what the model recognizes the patterns from these charts precisely without explainability.

Furthermore, the deep learning based models have several disadvantages, including lack of explanation capability~\cite{samek2017explainable} and difficulty in designing models. These difficulties will hinder the widely application of deep learning methodologies in critical fields. In this paper, we provide a framework based on adversarial attacks to demonstrate the adaptiveness and robustness of our model.

\subsection{Candlestick Pattern}
Following the chart is drawn from historical prices according to specific rules. These features help traders to see the price trend. The three more common types of charts are histograms, line charts, and the most widely used candlestick.

\begin{figure}[h]
\centering
\includegraphics[width=0.4\textwidth]{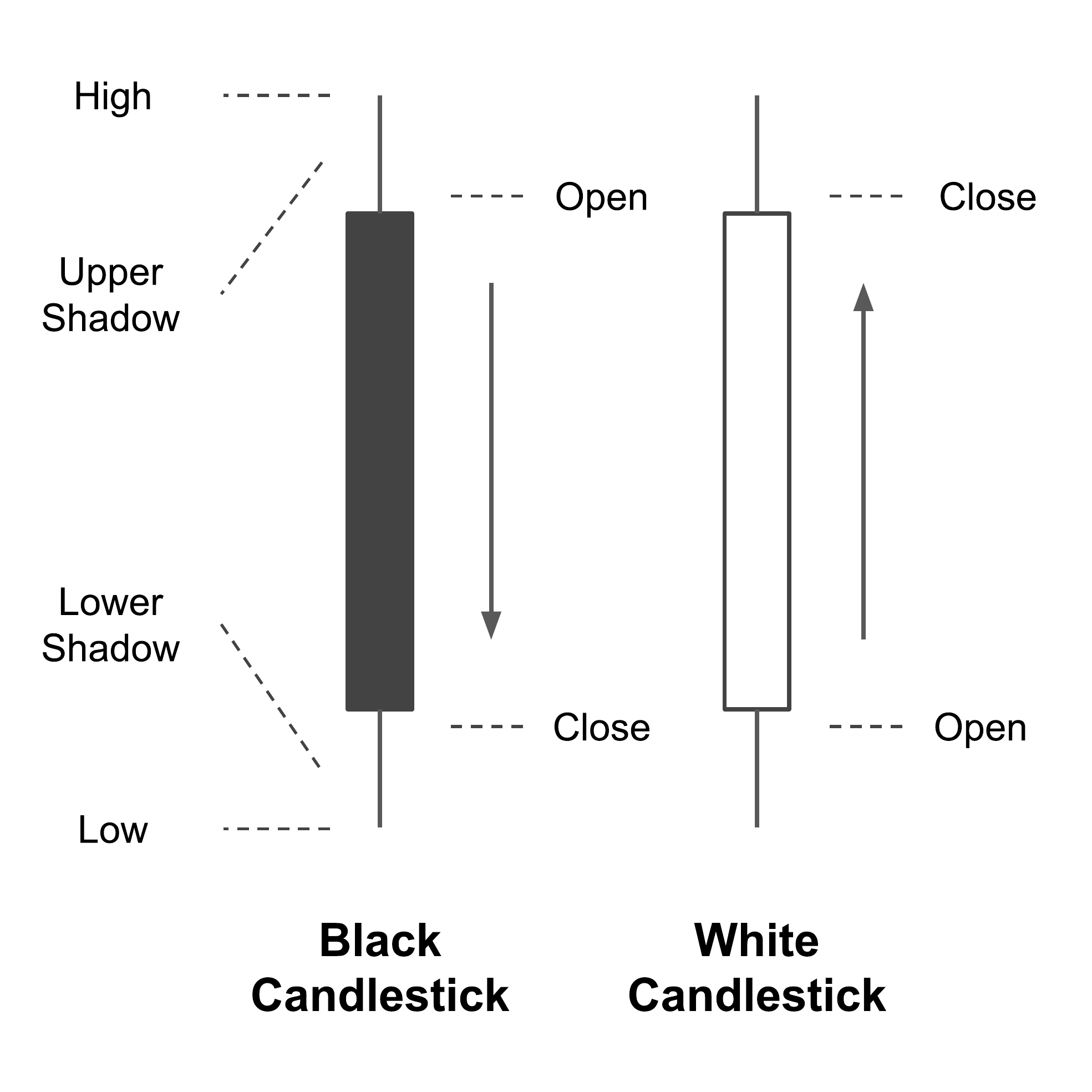}
\caption{The shape of a candlestick.}
\label{fig1}
\end{figure}

The candlestick is originated from Japan in the 17th century and has been popular in Europe and the United States for more than a century, especially in the foreign exchange market. As the most popular chart in technical analysis, traders should have an understanding of it.
It is named after a candle, as shown in Figure~\ref{fig1}.
A candlestick draws from the highest price, the lowest price, the opening price, and the closing price as follows:
\begin{enumerate}
\item Opening price: This price is the first price that occurs during the period;
\item Highest price: the highest price that occurs during the period;
\item Lowest price: the lowest price that occurs during the period;
\item Closing price: The last price that occurs during the period.
\end{enumerate}
If the closing price is higher than the opening price, the candlestick follows as:
\begin{enumerate}
\item the top of the candle body is the closing price;
\item the bottom is the opening price;
\item the color is usually green or white.
\end{enumerate}
If the closing price is lower than the opening price, the candlestick follows as:
\begin{enumerate}
\item the opening price above the candle body;
\item the closing price below;
\item the color is usually red or black.
\end{enumerate}

In some cases, the candlestick has no hatching because the opening or closing price coincides with the highest or lowest price. If the candle is very short, the opening and closing prices of the candlestick are very similar.

\subsection{Time Series System}
The candlestick draws in a coordinate system, with the horizontal axis representing time and the vertical axis representing price. Time is from left to right on the X-axis. The nearest candlestick is instantaneous within the corresponding period. Price is from top to bottom on the Y-axis. The higher the position of the candlestick, the higher the price in those markets at the time. Conversely, the lower the position of the candlestick, the lower the market price at that time.

There are a variety of chart cycles to choose from M1 (1-minute), M5 (5-minutes), M15 (15-minutes), M30 (30-minutes), H1 (1-hour), H4 (4-hours), D1 (1-day), and  W1 (1-week). The time required to form a candlestick calls the chart cycle. For example, the opening, closing, highest, and lowest price in 5 minutes plot into a candlestick. The candlestick composed of these prices is a 5-minute candlestick. Each candlestick represents a 5-minute price change.

\subsection{The 8 Most Powerful Candlestick Patterns}\label{8most}
\begin{figure}[h]
\centering
\includegraphics[width=0.45\textwidth]{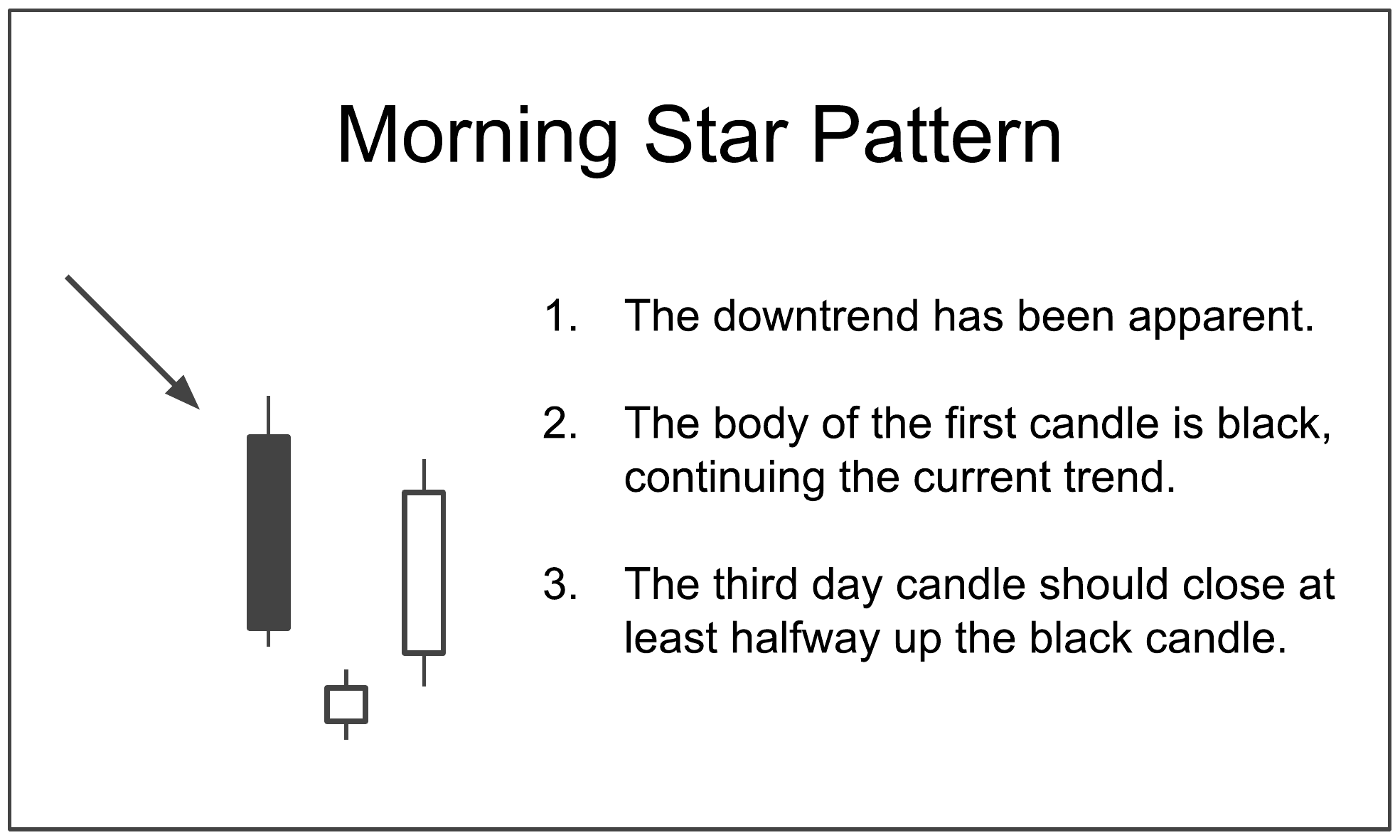}
\caption{{\bfseries Illustration of Morning Star Pattern.} The left-hand side shows the appearance of the Morning Star pattern. The right-hand side shows the critical rules of the Morning Star pattern.}
\label{morning_intro}
\end{figure}

\begin{figure}[h]
\centering
\includegraphics[width=0.45\textwidth]{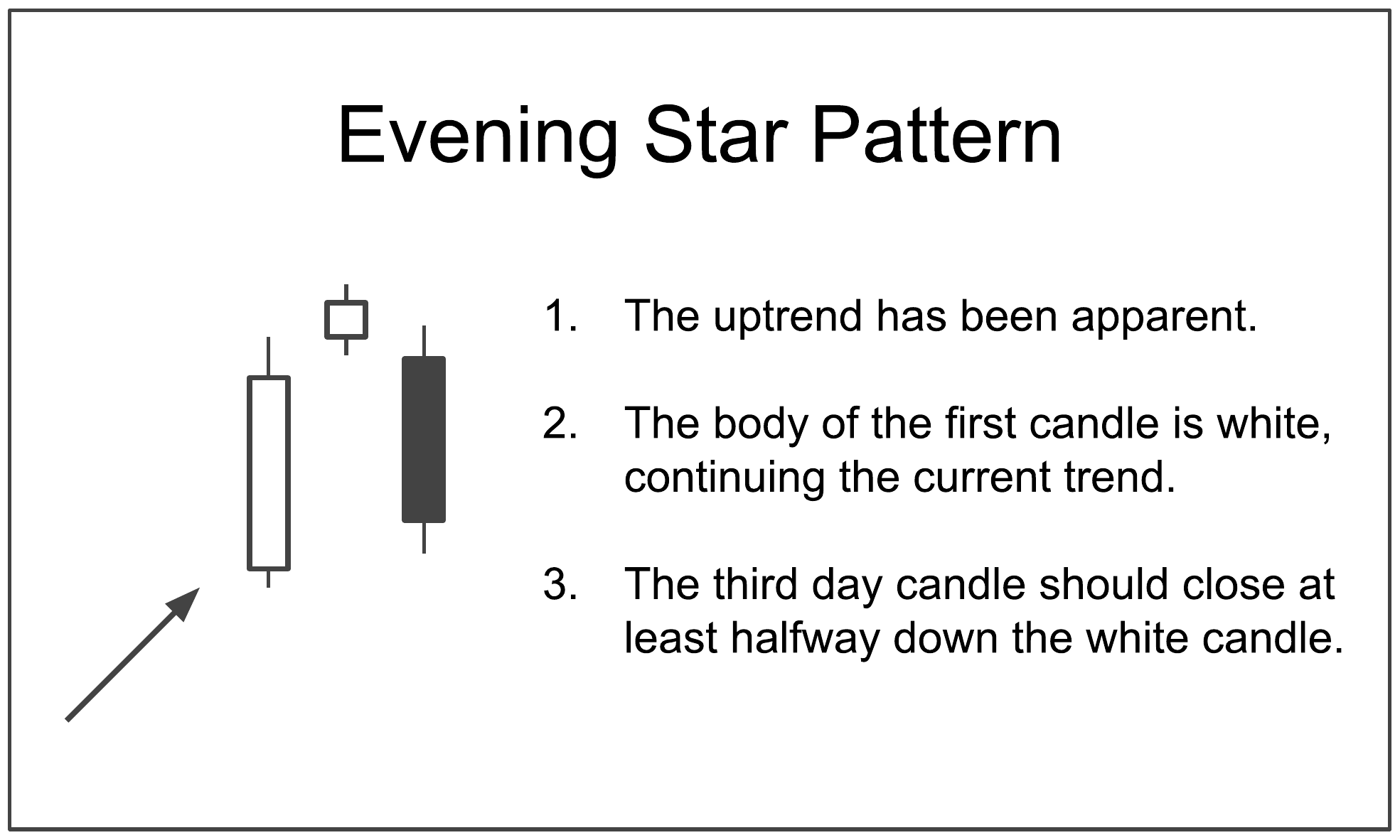}
\caption{{\bfseries Illustration of Evening Star Pattern.} The left-hand side shows the appearance of the Evening Star pattern. The right-hand side shows the critical rules of the Evening Star pattern.}
\label{evening_intro}
\end{figure}

The trick is in identifying some commonly occurring candlestick patterns and then building a market context around it. In the paper, we provide the most eight common candlestick patterns to analysis our explainable model as follows:
\begin{enumerate}
\item Morning Star is a visual pattern made up of a tall black candlestick, a smaller black or white candlestick with a short body and long wicks, and a third tall white candlestick. The middle candle of the morning star captures a moment of market indecision where the bears begin to give way to bulls. The third candle confirms the reversal and can mark a new uptrend. Figure~\ref{morning_intro} shows the morning star based on the description.
\item Evening Star is a bearish candlestick pattern consisting of the latest three candles: a large white candlestick, a small-bodied candle, and a black candle. The pattern will be more visible with a large black candle. Figure~\ref{evening_intro} shows the evening star based on the description.
\item Hammer occurs forms a hammer-shaped candlestick, in which the lower shadow is at least twice the size of the real body. The body of the candlestick represents the difference between the open and closing prices, while the shadow shows the high and low prices for the period.
\item Inverted Hammer looks like an upside down version of the hammer candlestick pattern, and when it appears in an uptrend is called a shooting star.
\item Bullish Engulfing forms when a small black candlestick is followed the next candle by a large white candlestick, the body of which completely overlaps or engulfs the body of the previous candlestick.
\item Bearish Engulfing consists of an up white candlestick followed by a large down black candlestick that eclipses or "engulfs" the smaller up candle.
\item Shooting Star is a bearish candlestick with a long upper shadow, little or no lower shadow, and a small real body near the low of the day. It appears after an uptrend.
\item Hanging Man is composed of a small real body, a long lower shadow, and little or no upper shadow. The hanging man shows that selling interest is starting to increase.
\end{enumerate}

\subsection{Explain our Model}
We propose a Gramian Angular Summation Field (GASF) time series encoding method to emphasize the time series features for the Convolutional Neural Networks (CNN) model.
The paper define the approach as a GASF-CNN model. Initially, we trained a GASF-CNN model that can capture the 8 major candlestick patterns with 90.0\% accuracy. Even though, we would like to know if the model learned the features as human seen. Hence, we used the Local Search Attack Adversarial model~\cite{8014906} to attack the GASF matrices. The attacked regions are on the main diagonal of the GASF matrices.

\begin{figure}[h]
\graphicspath{{figures}}
\begin{center}
\includegraphics[scale=0.4]{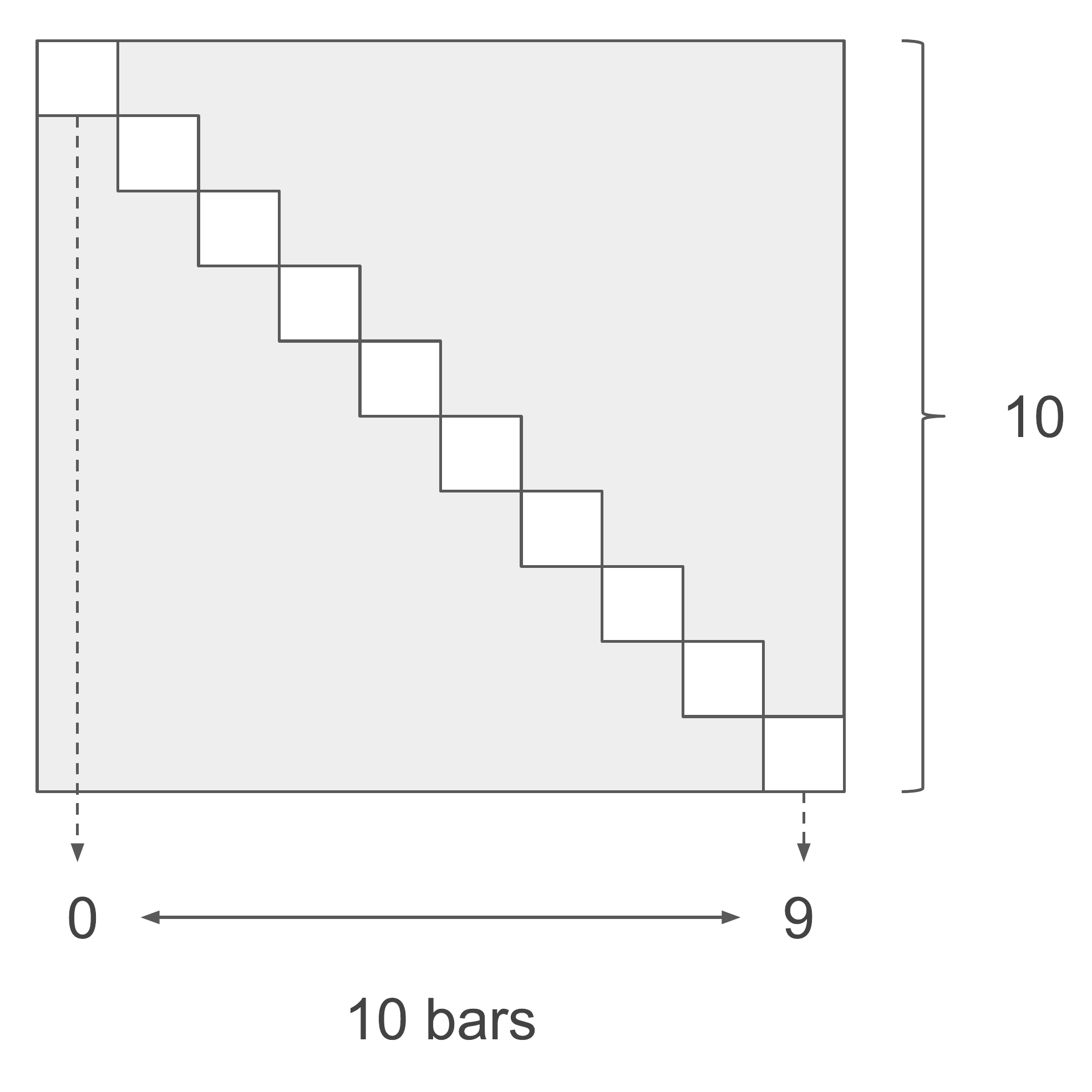}
\caption{The GASF matrix for the local search attack. In this work, we attack the diagonal elements as these locations }
\label{fig:attack_region}
\end{center}
\end{figure}

We defined these 10 bars based on the rules. The last 3 bars form the OHLC patterns, and a trend emerges in the rest of the bars. Figure~\ref{fig:attack_region} illustrates the attack region on GASF matrix.

1500 GASF matrices were attacked in each label. If most of the GASF matrices can be attacked successfully, it means the region human seen is similar to what the GASF-CNN model has learned.

\section{Methods}
\subsection{GASF-CNN}
A two-step approach is adopted, Gramian Angular Summation Field (GASF) time series encoding~\cite{wang2015encoding} and Convolutional Neural Networks (CNN) model.

First, we encode time series data based on open, high, low, and close prices to GASF matrices. GASF is a novel time series encoding method proposed by Wang and Oates, which represents time series data in a polar coordinate system and uses various operations to convert these angles into a symmetry matrix.

The first step to make a GASF matrix is to normalize the time series data into [0, 1]. Secondly, represent the normalized time series data in the polar coordinate system. Equation \ref{gasf:1} and \ref{gasf:2} show how to get the angles and radius from the rescaled time series data. Finally, we sum the angles and use the cosine function to make the GASF as equation \ref{gasf:3}.

\begin{align} 
\phi&=\arccos(\widetilde{x}_i), -1\leq \widetilde{x}_i \leq 1, \widetilde{x}_i \in \widetilde{X}\label{gasf:1}\\
r&=\frac{t_i}{N}, t_i\in\mathbb{N}\label{gasf:2}\\
\textup{GASF} &=\cos(\phi_i + \phi_j) \\ &=
\widetilde{X}^T \cdot \widetilde{X} - \sqrt{I-\widetilde{X}^2}^T\cdot \sqrt{I-\widetilde{X}^2}\label{gasf:3}
\end{align}

Once GASF transform is completed, the 3-d data are able to be inputs for the CNN model training.
The architecture of our CNN model is similar to LeNet~\cite{lecun2015lenet}, including two convolutional layers with 16 kernels and one fully-connected layer with 128 denses.

\subsection{Local Search Attack}
In this work, we apply the following scheme to investigate the possible regions that are critical for the classification process. Logically, if a pixel is important in the final classification result, then a perturbation of that pixel should result in a degradation of the confidence score or even a misclassification. To achieve of this, we propose a method which is modified from the \emph{local search attack}~\cite{8014906}. First of all, we define the set of points that can be perturbed. In this work, in order to maintain the consistency of the original time-series data and the GASF matrix, we only perturbate the diagonal elements in the GASF matrix. Once we obtain the perturbated diagonal elements, we then calculate the corresponding values of non-diagonal elements and output the perturbated GASF matrix.
Secondly, send this perturbated GASF into the CNN model to get the classification results. If the perturbated input is not misclassified, simply repeat the procedure described above. The detail of the algorithm is in Algorithm~\ref{alg}.

\begin{algorithm}[tb]
\begin{algorithmic}
\State Load a single GASF two-dimensional array $A$
\State Set $T = \text{length of the time-series}$ 
\State Keep a copy of $A$ in memory $D$
\State Initialize the counter $t = 0$
\For{episode $=1,2,\ldots,R$}

\If{$t = 10$}
\State Reinitialize the $A$ to the original value from memory $D$
\State Reset the counter $t = 0$
\EndIf
\For {$l = 1,2,\ldots,T$}
	\State Sampling a random perturbation scale $r_l$ from uniform distribution $[0.8,1.2]$
	\State Calculate the perturbated result $=r_l \times A[l,l]$
	\If{$r_l \times A[l,l] \geq 0.5 \lor r_l \times A[l,l] \leq -0.5$}
	\State $A[l,l]$ keeps the original value.
	\Else
	\State Set $A[l,l] = r_l \times A[l,l]$ 
	\EndIf
\EndFor

\State $t = t + 1$
\State Recalculate the time-series from perturbated $A$ and then encode into a new GASF matrix $A'$
\If{$A'$ \text{is adversarial}}
\algorithmicreturn{$A'$}
\EndIf
\EndFor
\end{algorithmic}
\caption{Local Search Attack}
\label{alg}
\end{algorithm}

\begin{table}[]
\centering
\begin{tabular}{|c|c|}
\hline
Parameters & Values            \\ \hline
r          & uniform(0.8, 1.2) \\ \hline
d          & 0                 \\ \hline
t          & 10                \\ \hline
R          & 150               \\ \hline
reset      & 10                \\ \hline
\end{tabular}
\caption{The parameters of our local search attack model.}
\label{tab:attack_parameters}
\end{table}

\section{Results}
\begin{figure}[h]
\begin{center}
\subfigure[The first attack result example.]{\includegraphics[width=1.0\linewidth, height= 4cm]{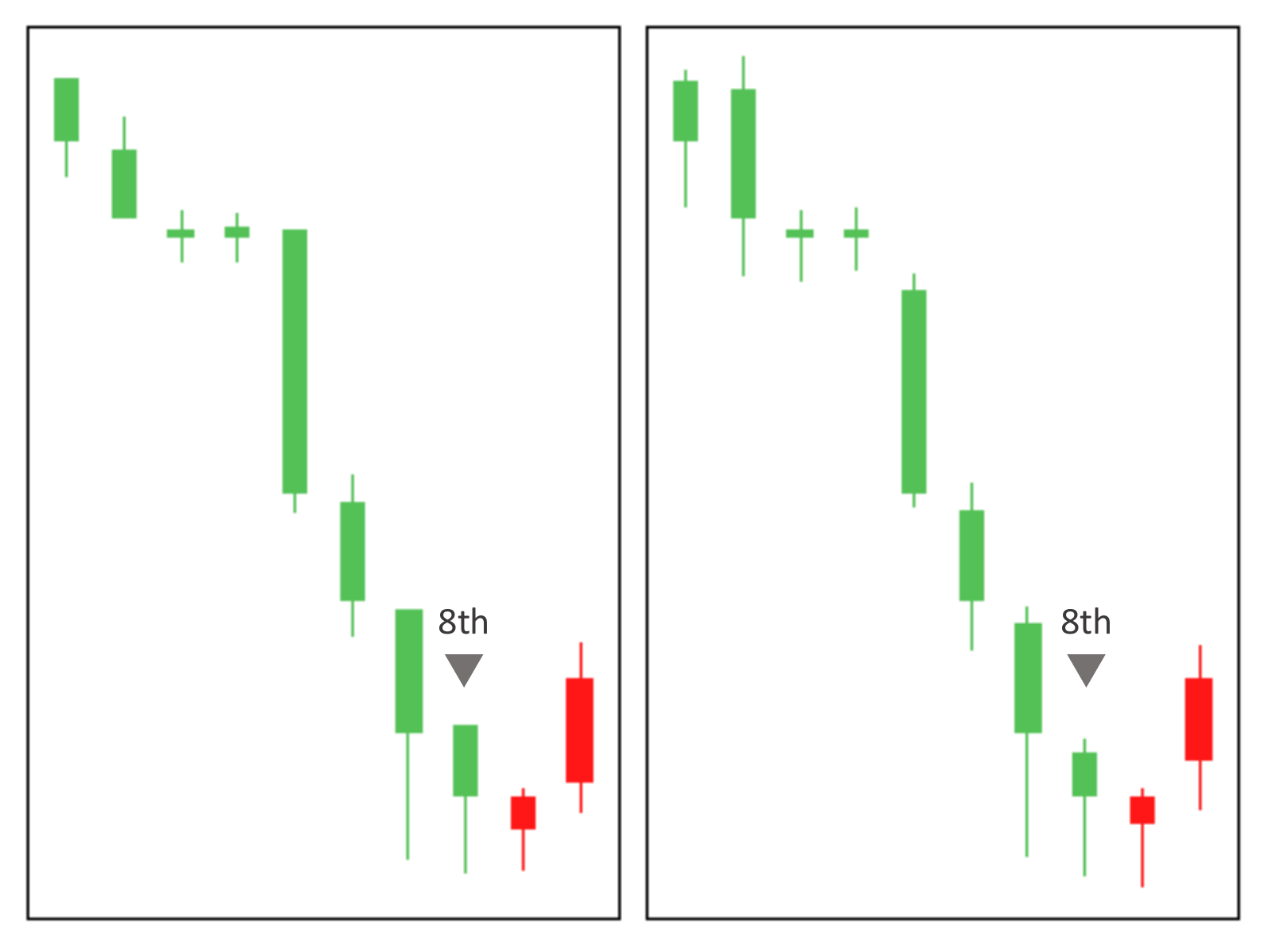}}
\subfigure[The second attack result example.]{\includegraphics[width=1.0\linewidth, height= 4cm ]{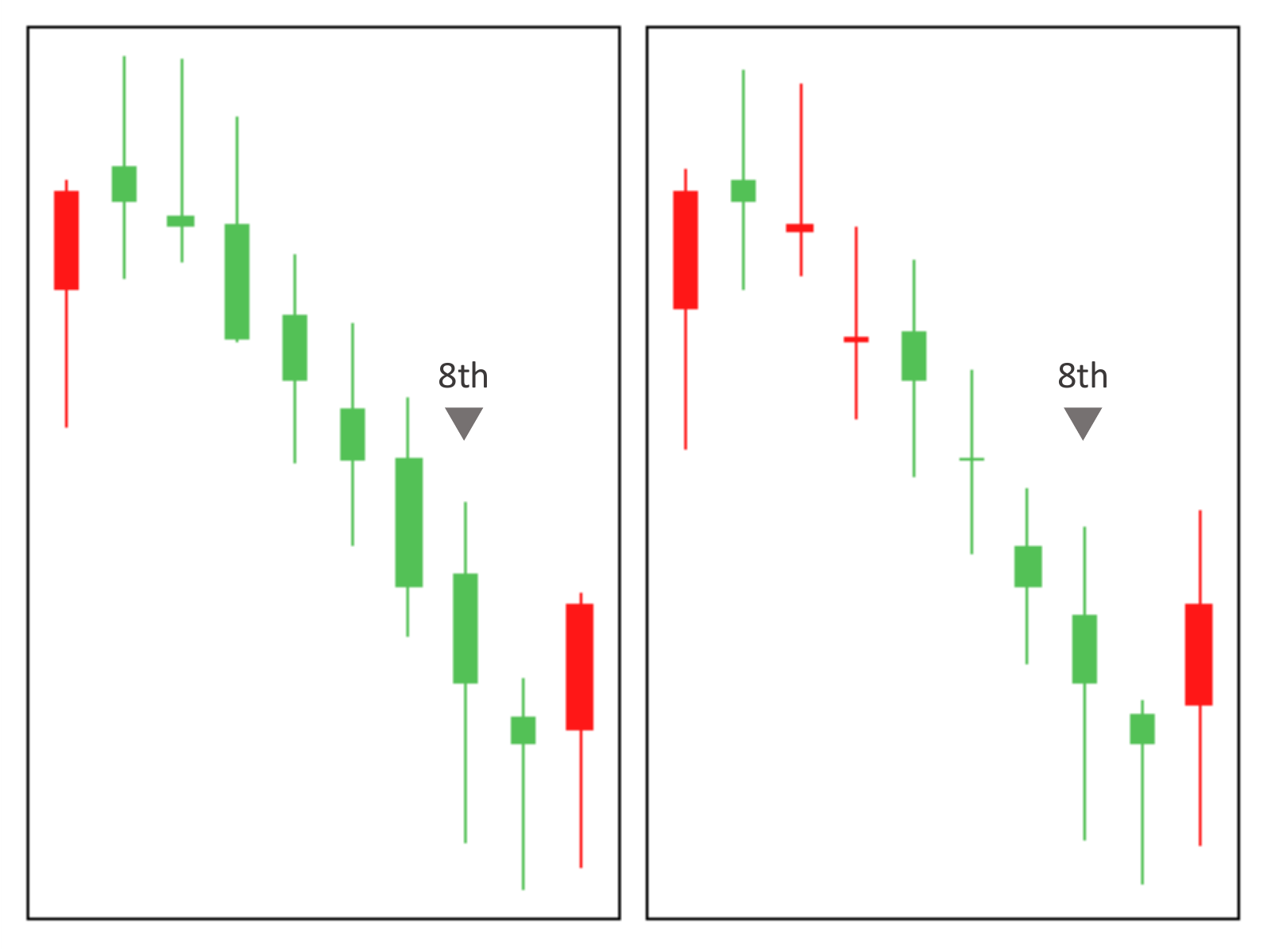}}
\end{center}
\caption{The attack result example of morning star pattern.}
\label{fig:morning_r8}
\end{figure}

\begin{figure}[h]
\begin{center}
\subfigure[The first attack result example.]{\includegraphics[width=1.0\linewidth, height= 4cm]{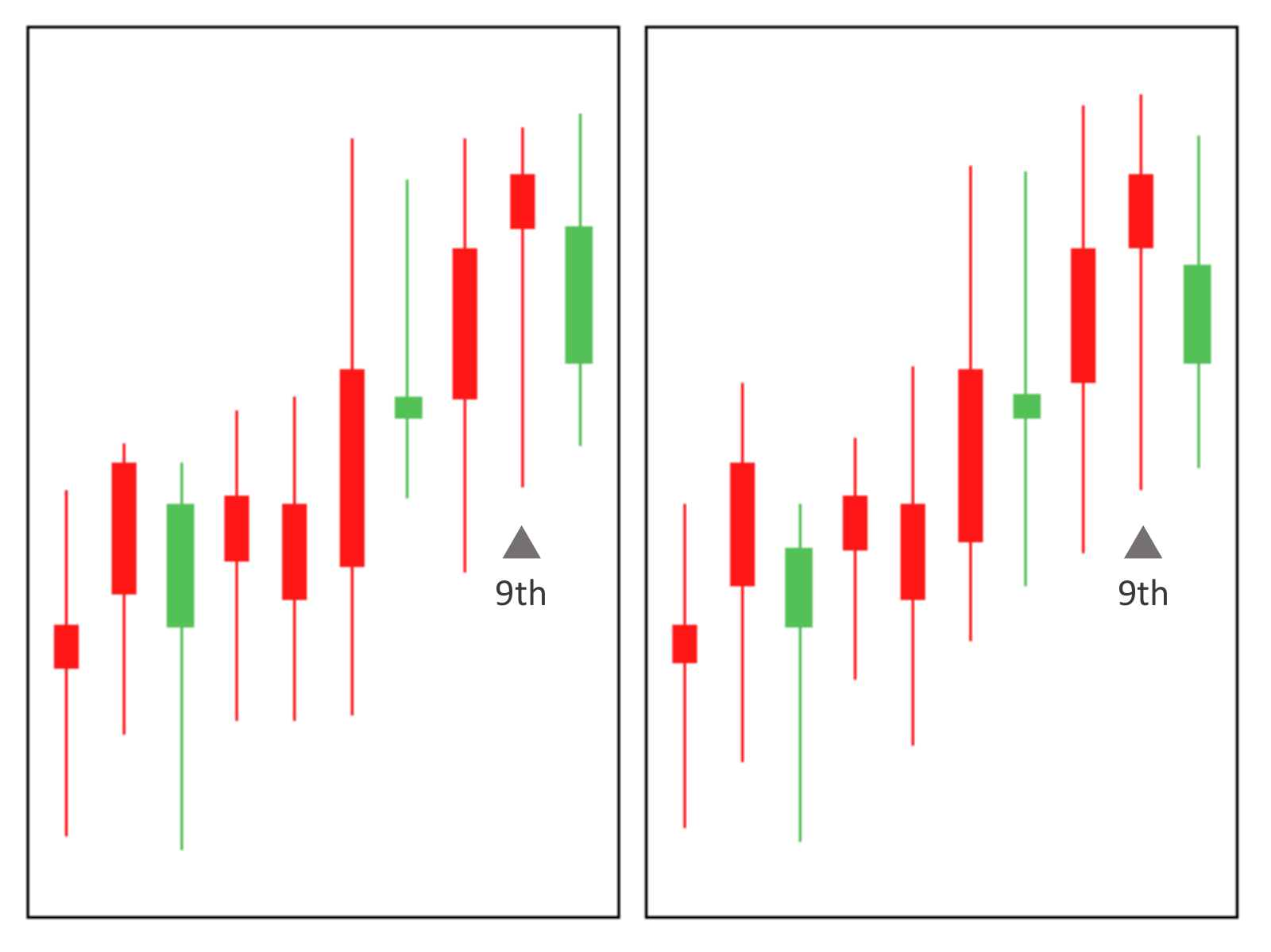}}
\subfigure[The second attack result example.]{\includegraphics[width=1.0\linewidth, height= 4cm ]{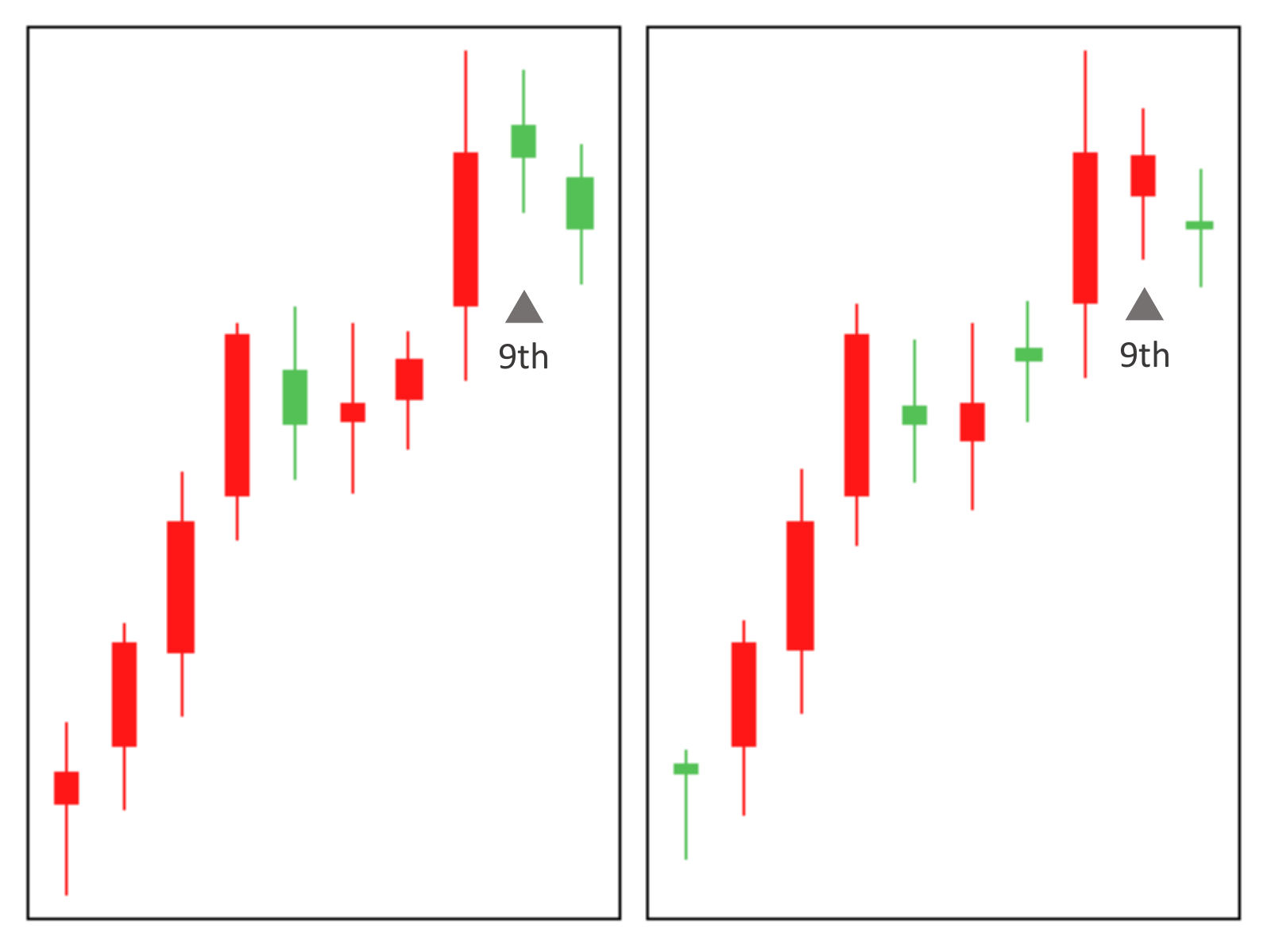}}
\end{center}
\caption{The attack result example of evening star pattern.}
\label{fig:evening_r8}
\end{figure}

\begin{table}[]
\centering
\begin{tabular}{|c|c|c|}
\hline
Label & Success Rate & Percent (\%) \\ \hline
1     & 631 / 1500       & 42.1         \\ \hline
2     & 972 / 1500       & 64.8         \\ \hline
3     & 1079 / 1500      & 71.9         \\ \hline
4     & 1319 / 1500      & 87.9         \\ \hline
5     & 602 / 1500       & 40.1         \\ \hline
6     & 932 / 1500       & 62.1         \\ \hline
7     & 953 / 1500       & 63.5         \\ \hline
8     & 1238 / 1500      & 82.5         \\ \hline
\end{tabular}
\caption{The attack ratio of local search attack for each label.}
\label{tab:attack_result}
\end{table}
We use EUR/USD 1-minute open, high, low, and close price data to produce our empirical results. The training data is from January 1, 2010 to January 1, 2016. The testing data is from January 2, 2016 to January 1, 2018. There are eight patterns and each label includes 1500 data. If the pattern does not belong to any one of the eight patterns, we set that kind of patterns as the label 0 and there are 3000 samples in this category. All of the data can be downloaded from URL:~\url{https://github.com/pecu/FinancialVision}.

These data produce the following results.
Figure~\ref{fig:morning_r8} shows the result of attacked morning star pattern.
Figure~\ref{fig:evening_r8} shows the result of attacked evening star pattern. 
With the modified local search attack model, we can reach 64.36\% success attack rate on average. Table~\ref{tab:attack_result} presents the full results with at least 40.0\%. This result suggests that it's plausible to focus attack region on the diagonal. Our GASF-CNN model actually recognizes the diagonal patterns, where the last 3 bars form the major patterns and the rest represent the trend.

\section{Discussion}
We use two of the eight patterns for discussion. According to the description of the subsection~\ref{8most}, the morning star pattern composes of downtrend and three-bar pattern: a large black candlestick, a small-bodied candle, and a white candle. After the perturbation, both Figure~\ref{fig:morning_r8}-(a) and Figure~\ref{fig:morning_r8}-(b) have no significant change, but the 8th and the last bar become small enough to make whole pattern invisible. The left hand side shows the original pattern, and the right hand side shows the pattern after attack.

The evening star pattern composes of uptrend and three-bar pattern: a large white candlestick, a small-bodied candle, and a black candle. After the perturbation, the three-bar pattern violate the rules. In both Figure~\ref{fig:evening_r8}-(a) and Figure~\ref{fig:evening_r8}-(b), the 9th bar become larger, which is too big for evening star, and the last bar become smaller making the whole pattern invisible. The left hand side shows the original pattern, and the right hand side shows the pattern after attack. 

The two empirical results show that our local search adversarial attack approach can explain the GASF-CNN model learned as human has seen. We can clearly understand how GASF-CNN model recognize candlestick pattern. 
Hence, our explainable GASF-CNN model is much more trustworthy and reliable for traders compared to others without knowing the underlying learning experience. 

\section{Conclusion}
This paper has two main contributions. The first is that the paper proposes a GASF-CNN model to construct an innovation field of financial vision research for candlestick recognition. The second is that the paper proposes an approach based on modified local search adversarial attack to explain the reason for the GASF-CNN model on how to determine the different candlestick patterns.

The GASF-CNN model can identify the eight types of candlestick. According to our results of the local search adversarial attacks, the GASF-CNN model understands the feeling of the candlestick as a human has seen. If we can confirm through the analytical method that the GASF-CNN model has indeed learned the sense of the candlestick from the trader, then it will be perfect for building a complete explainable trading model. If the last piece of the feeling of the candlestick in monitoring tool supplements in the future, it will expand new trading research fields.

We provide a open source implementation for the local search adversarial attack approach in the follow URL:~\url{https://github.com/pecu/FinancialVision}.

\bibliographystyle{named}
\bibliography{ijcai20}

\begin{thebibliography}{}

\bibitem[\protect\citeauthoryear{LeCun and others}{2015}]{lecun2015lenet}
Yann LeCun et~al.
\newblock Lenet-5, convolutional neural networks.
\newblock {\em URL: http://yann. lecun. com/exdb/lenet}, page~20, 2015.

\bibitem[\protect\citeauthoryear{{Narodytska} and
  {Kasiviswanathan}}{2017}]{8014906}
N.~{Narodytska} and S.~{Kasiviswanathan}.
\newblock Simple black-box adversarial attacks on deep neural networks.
\newblock In {\em 2017 IEEE Conference on Computer Vision and Pattern
  Recognition Workshops (CVPRW)}, pages 1310--1318, July 2017.

\bibitem[\protect\citeauthoryear{Samek \bgroup \em et al.\egroup
  }{2017}]{samek2017explainable}
Wojciech Samek, Thomas Wiegand, and Klaus-Robert M{\"u}ller.
\newblock Explainable artificial intelligence: Understanding, visualizing and
  interpreting deep learning models.
\newblock {\em arXiv preprint arXiv:1708.08296}, 2017.

\bibitem[\protect\citeauthoryear{Tsai \bgroup \em et al.\egroup
  }{2019}]{tsai2019encoding}
Yun-Cheng Tsai, Jun-Hao Chen, and Chun-Chieh Wang.
\newblock Encoding candlesticks as images for patterns classification using
  convolutional neural networks.
\newblock {\em arXiv preprint arXiv:1901.05237}, 2019.

\bibitem[\protect\citeauthoryear{Wang and Oates}{2015}]{wang2015encoding}
Zhiguang Wang and Tim Oates.
\newblock Encoding time series as images for visual inspection and
  classification using tiled convolutional neural networks.
\newblock In {\em Proceedings of the 2015 Association for the Advancement of
  Artificial Intelligence (AAAI) Workshops}, pages 40--46, 2015.

\end{thebibliography}

\end{document}